# Feature Quality and Adaptability of Medical Foundation Models: A Comparative Evaluation for Radiographic Classification and Segmentation

Frank Li[1], Theo Dapamede[1], Mohammadreza Chavoshi[1], Young Seok Jeon[1], Bardia Khosravi[2], Abdulhameed Dere[3], Beatrice Brown-Mulry[1], Rohan Satya Isaac[1], Aawez Mansuri[1], Chiratidzo Sanyika[1], Janice Newsome[1], Saptarshi Purkayastha[4], Imon Banerjee[5], Hari Trivedi[1], Judy Gichoya[1]

[1]Department of Radiology, Emory University, Atlanta, GA, USA
[2]Department of Radiology, Mayo Clinic, Rochester, MN, USA
[3]Faculty of Clinical Sciences, College of Health Sciences, University of Ilorin, Ilorin, Nigeria
[4]Luddy School of Informatics, Computing, and Engineering, Indiana University Indianapolis, Indianapolis, IN, USA
[5]Department of Radiology, Mayo Clinic, Phoenix, AZ, USA

**Corresponding Author:**
Judy Gichoya
judywawira@emory.edu

## ABSTRACT

Foundation models (FMs) promise to generalize medical imaging, but their effectiveness varies. It remains unclear how pre-training domain (medical vs. general), paradigm (e.g., text-guided), and architecture influence embedding quality, hindering the selection of optimal encoders for specific radiology tasks.

To address this, we evaluate vision encoders from eight medical and general-domain FMs for chest X-ray analysis. We benchmark classification (pneumothorax, cardiomegaly) and segmentation (pneumothorax, cardiac boundary) using linear probing and fine-tuning.

Our results show that domain-specific pre-training provides a significant advantage; medical FMs consistently outperformed general-domain models in linear probing, establishing superior initial feature quality. However, feature utility is highly task-dependent. Pre-trained embeddings were strong for global classification and segmenting salient anatomy (e.g., heart). In contrast, for segmenting complex, subtle pathologies (e.g., pneumothorax), all FMs performed poorly without significant fine-tuning, revealing a critical gap in localizing subtle disease.

Subgroup analysis showed FMs use confounding shortcuts (e.g., chest tubes for pneumothorax) for classification, a strategy that fails for precise segmentation. We also found that expensive text-image alignment is not a prerequisite; image-only (RAD-DINO) and label-supervised (Ark+) FMs were among top performers. Notably, a supervised, end-to-end baseline remained highly competitive, matching or exceeding the best FMs on segmentation tasks.

These findings show that while medical pre-training is beneficial, architectural choices (e.g., multi-scale) are critical, and pre-trained features are not universally effective, especially for complex localization tasks where supervised models remain a strong alternative.

# INTRODUCTION

Recent years have witnessed the rapid development of deep learning models for medical imaging, often achieving performance that rivals or surpasses human experts (1). Despite their success, these models, typically trained with supervised learning require massive meticulously annotated datasets and are designed for narrow predictive tasks. Furthermore, they are overfit the narrow training data distribution and can exhibit performance disparities on external populations, including underrepresented groups (2).

Foundation models (FMs), pretrained on large heterogenous datasets using self-supervised techniques, represent a paradigm shift aimed at overcoming some of these challenges. In contrast to traditional deep learning models, characterized by their ability to generalize with zero-shot, FMs can also be fine-tuned for a wide range of downstream applications using significantly less data thereby reducing the time and cost of model training and data annotation for new use cases (3). Among FMs, vision-language models (VLMs) have emerged as a promising category due to their ability to integrate visual and textual information (4). A typical VLM consists of a vision encoder that extracts features from images and a text decoder that processes these features alongside user prompts to generate human-readable text. VLMs enable intuitive, natural language interaction for tasks like disease diagnosis, report generation, and phrase grounding (5). Beyond their generative capabilities, the vision encoders of these models are powerful feature extractors. These learned features, or embeddings, can be used to train lightweight models (i.e., adapters) for various downstream tasks through methods like linear probing (training a classifier on frozen features) or fine-tuning (partial update of the encoder's weights) (6). However, the quality of these embeddings is heavily dependent on the encoder's pre-training strategy, leading to a diverse and complex landscape of available models.

Two major pre-training strategies have emerged for medical vision encoders. The most common is Text-guided Contrastive Learning, which uses methods like CLIP (7) to align image and text embeddings in a shared latent space. This approach has been adapted for the medical domain in numerous FMs, including BiomedCLIP (8), CheXagent (9), MedImageInsight (10), and MedSigLIP (11), all of which learn visual features from paired image-text data. This strategy's primary limitation is its dependency on the quality of the accompanying text; incomplete or generic radiological reports can lead to "representation collapse (12)", where the model oversimplifies visual features to match the text, failing to capture subtle but clinically significant intra-class variations. A second strategy, Image-only Self-Supervised Learning, challenges this reliance on text. Exemplified by models like RAD-DINO (12), which adapts the DINOv2 (13) image-only self-supervised learning approach, this approach pre-trains the image encoder solely on imaging data, learning rich visual features directly from pixels and avoiding the constraints of textual supervision. An adapter can be then trained to project the frozen image encoder's embeddings onto the language space and then fine-tuned with the text decoder. Beyond these VLM-specific approaches, supervised Multi-Source Learning (14), represents an alternative direction in medical imaging FMs. Demonstrated by models like Ark+ (15) employs a teacher-student framework to cyclically accrue expert knowledge from numerous public datasets with heterogeneous labels, aggregating diverse expertise to build a robust and generalizable model. Though Ark+ operates outside the VLM paradigm, it exemplifies how alternative pre-training strategies continue to evolve in medical imaging, offering a different path to building generalizable foundation models.

These distinct pre-training paradigms are further complicated by significant variations in training data and model architecture, which act as confounding variables. For instance, data scale and diversity range widely from radiology-specific datasets like the 838,000 images used for RAD-DINO, to broad, multi-modal biomedical collections like the 15 million image-text pairs from scientific articles used for BiomedCLIP, and the 14 medical imaging domains used for MedImageInsight. Architectural choices also differ, with models employing various backbones such as Swin-Large (16) for Ark+, ViT-Base (17) for RAD-DINO and BiomedCLIP, and DaViT (18) for MedImageInsight. These differences in data and architecture make direct, "apples-to-apples" comparisons of pre-training strategies challenging. **Table 1** shows a summary of the models' specifications.

**Table 1.** Overview of Evaluated FMs. This table summarizes the key characteristics of the vision FMs used in this study, including both medical domain-specific and general-purpose architectures, with SegFormer serving as the baseline benchmark.

| Model | Year | Multi-imaging Modal | Image Encoder | Pretraining Method | Text Guided | #Parameters | Input Image Size | Embedding Size Global | Embedding Size Patch (C*H*W) |
|---|---|---|---|---|---|---|---|---|---|
| Ark+ | 2025 | No | Swin-Large | Cyclical Knowledge Accumulation and Reuse (Supervised) | No | 195M | 768*768 | 1536 | 1536*24*24, 768*48*48, 384*96*96, 192*192*192 |
| BiomedCLIP | 2023 | Yes | ViT-B/16 | CLIP (Self-supervised) | Yes | 195M | 224*224 | 512 | 768*14*14 |
| CheXagent* | 2024 | No | SigLIP-Large | SigLIP (Self-supervised) | Yes | 316M | 512*512 | 1024 | 1024*32*32 |
| MedImageInsights | 2024 | Yes | DaViT | UniCL (Self-supervised) | Yes | 360M | 480*480 | 1024 | 2048*15*15, 1024*30*30, 512*60*60, 256*120*120 |
| MedSigLIP | 2025 | Yes | SigLIP2-So400m | SigLIP (Self-supervised) | Yes | 428M | 448*448 | 1152 | 1152*32*32 |
| RAD-DINO | 2024 | No | ViT-B/14 | MIM, SSID (Self-supervised) | No | 86.6M | 518*518 | 768 | 768*37*37 |
| DINOv2 | 2023 | No | ViT-B/14 | MIM, SSID (Self-supervised) | No | 86.6M | 224*224 | 768 | 768*16*16 |
| SigLIP2 | 2025 | No | SigLIP2-So400m | SigLIP (Self-supervised) | Yes | 428M | 512*512 | 1152 | 1152*32*32 |
| SegFormer | 2021 | No | - | - | - | 64M | 512*512 | - | - |

MIM: Masked Image Modeling; SSID: Self-Supervised Instance Discrimination; CLIP: Contrastive Language-Image Pre-Training; UniCL: Unified Contrastive Learning in Image-Text-Label Space; SigLIP: Sigmoid Loss for Language Image Pre-training; C: Number of channels; H: Height of the image; W: Width of the image. *SIIM-ACR was used to pre-train the FM.

Despite rapid progress in multimodal foundation models, it remains unclear how differences in pre-training paradigms—compounded by variations in data composition and architectural design—affect the quality and adaptability of the learned embeddings. There is limited understanding of which vision encoders are best suited for specific downstream radiology tasks. Current evaluations typically emphasize classification accuracy, which may not adequately capture an encoder's capacity to represent fine-grained, localized anatomical or pathological features essential for clinical interpretation.

In this study, we address this gap by providing an evaluation of vision encoders from state-of-the-art medical FMs. We assess the quality and adaptability of their learned embeddings by evaluating performance on both classification and segmentation tasks for chest radiographs. By including segmentation, which demands a high degree of spatial and textural discrimination, we a probe into the fine-grained feature representation capabilities of each encoder. Our analysis aims to provide clear insights into the strengths and weaknesses of different pre-training strategies, guiding researchers and practitioners 6in selecting the most appropriate FMs for the downstream clinical applications.

## METHODS

Experimental Design

We designed a two-stage evaluation process for downstream classification and segmentation tasks on chest radiographs (**Figure 1**). For direct and fair comparison, the same linear head layer was used for a given task in both evaluation stages. First, to assess the inherent quality of the pretrained features, we employed linear probing (19). In this stage, we froze the weights of the model's vision encoder and trained only the attached linear prediction head. This approach measures how well the "off-the-shelf" features can solve a new task without any weight modification to the base model.

Second, to evaluate feature adaptability, we fine-tuned the entire model. Both the vision encoder and the prediction head were trained, but we used differential learning rates (20): a small rate (1e-7) for the vision encoder and a larger rate (1e-4) for the prediction head. This strategy allows the encoder to adapt to the new task while minimizing catastrophic forgetting of the powerful knowledge acquired during pre-training. For classification tasks, the prediction head was attached to the global embedding, which captures a holistic representation of the image. For segmentation, the head was attached to the unrolled patch embeddings, which provide the granular, spatial information necessary for pixel-level predictions.

For hierarchical models (Ark+ and MedImageInsights), which generate features at multiple scales, we conducted a preliminary analysis to identify the optimal feature resolution. We evaluated patch embeddings from two separate layers, corresponding to effective patch sizes of 16x16 and 32x32. The features from the 16x16 patch size consistently yielded superior segmentation performance and were therefore selected for all subsequent comparisons against the other models (**Figure S1**).

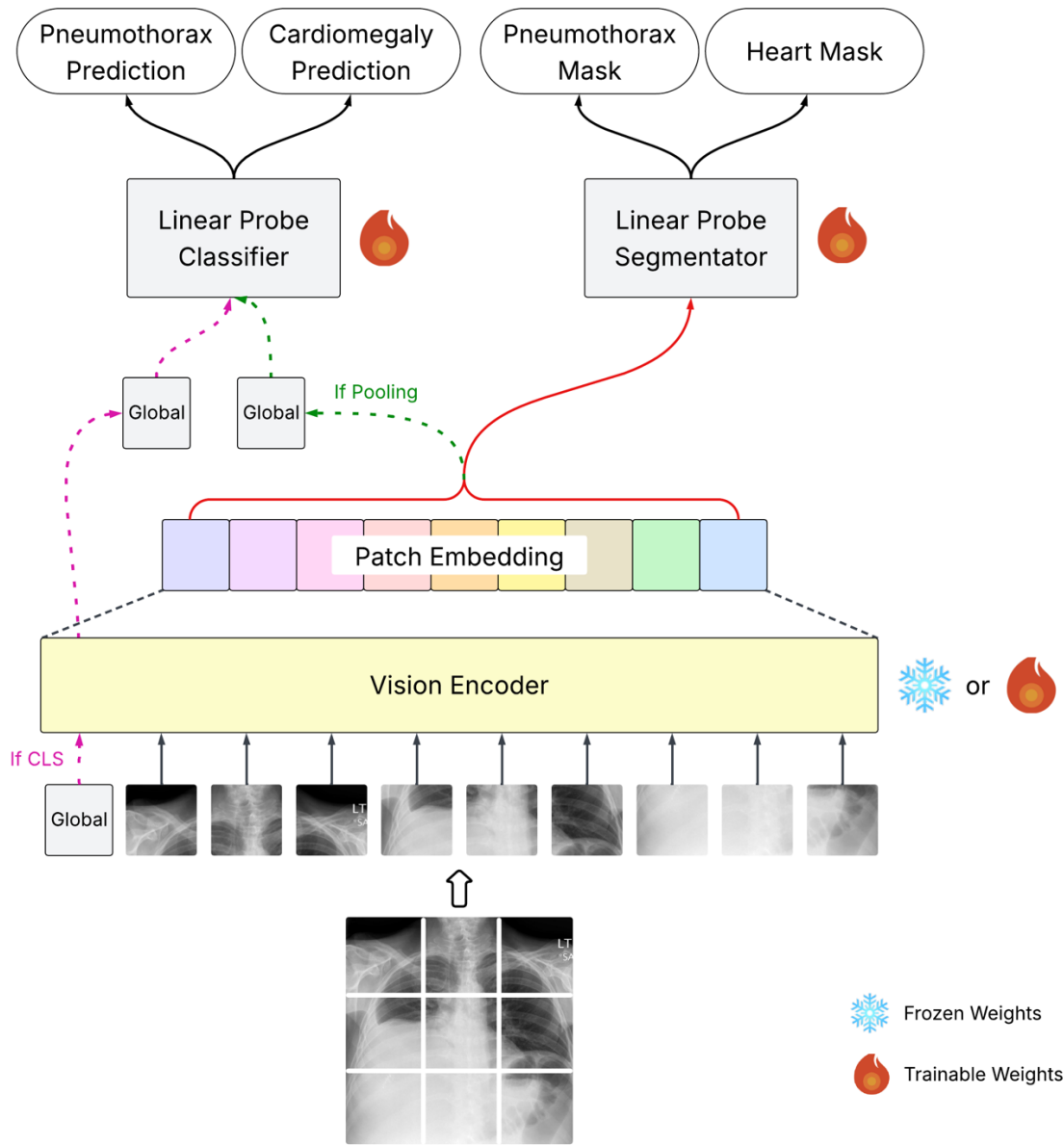

**Figure 1.** Evaluation framework for assessing vision encoder embeddings. Our two-phase approach first involves extracting features from chest X-ray images using the vision encoder of a pre-trained FM. For downstream classification tasks, such as predicting pneumothorax or cardiomegaly, the global token embedding is used to train a lightweight, trainable linear probe classifier. For segmentation tasks, like generating masks for the heart or pneumothorax, the per-patch embeddings are utilized to train a linear probe segmentator. Depending on the model's architecture, the global embedding is derived from either the output of a learnable [CLS] token or a pooling operation applied to the final patch embeddings. This framework accommodates two evaluation settings: linear probing, where the vision encoder's weights are frozen to assess the intrinsic quality of the embeddings, and fine-tuning, where the encoder's weights are updated to measure the model's adaptability.

Models and Datasets

We evaluated eight Transformer-based FMs. Six of these were medical-domain models pretrained on datasets that included chest X-rays (CXRs): Ark+, BiomedCLIP, CheXagent, MedImageInsights, MedSigLIP, and RAD-DINO. The other two were DINOv2 and SigLIP2, which are general domain models included to measure the benefit of domain-specific pretraining. Additionally, we trained SegFormer, a transformer-based model, from scratch to serve as a supervised baseline and benchmark the value of large-scale pretraining.

For training and evaluation, we used two datasets focused on common thoracic pathologies: pneumothorax and cardiomegaly. For pneumothorax, we used the SIIM-ACR pneumothorax dataset (21), which contains 10,675 CXRs. The classification subset was balanced (n=4,758) to mitigate prevalence bias (22), while the segmentation task used only positive cases. **Table 2** summarizes the cohort characteristics based on presence of chest tube and pneumothorax volume grouped into quartiles (Q1-Q4). For cardiomegaly, we used a balanced cohort of posterior-anterior CXRs from a private dataset from our institution (n=5,000, **Table 3**). All segmentation masks for the heart were generated using the CheXmask methodology (23). To ensure robust and generalizable results, we used a five-fold, patient-level cross-validation with a 64-16-20 split for training, validation, and testing across all tasks.

**Table 2.** Cohort Characteristics by Chest Tube Status and Pneumothorax Volume Quartiles (Q1-Q4)

|  |  | Overall | Absent | Pneumothorax Present | | | |
|---|---|---|---|---|---|---|---|
|  |  |  |  | Q1 | Q2 | Q3 | Q4 |
| n |  | 4758 | 2379 | 595 | 595 | 594 | 595 |
| ChestTube, n (%) | Absent | 2739 (57.6) | 2113 (88.8) | 138 (23.2) | 154 (25.9) | 161 (27.1) | 173 (29.1) |
|  | Present | 2019 (42.4) | 266 (11.2) | 457 (76.8) | 441 (74.1) | 433 (72.9) | 422 (70.9) |
| Volume%, mean (SD) |  | 1.3 (1.5) |  | 0.2 (0.1) | 0.6 (0.1) | 1.2 (0.3) | 3.3 (1.9) |

**Table 3.** Demographic Characteristics of the Study Cohort by Cardiomegaly Status

| | | Overall | Cardiomegaly Absent | Cardiomegaly Present |
|---|---|---|---|---|
| n | | 5000 | 2500 | 2500 |
| Sex, n (%) | F | 2651 (53.0) | 1319 (52.8) | 1332 (53.3) |
| | M | 2270 (45.4) | 1156 (46.2) | 1114 (44.6) |
| | NA | 79 (1.6) | 25 (1.0) | 54 (2.2) |
| Age, mean (SD) | | 56.5 (17.0) | 52.3 (17.3) | 60.7 (15.7) |
| BMI, mean (SD) | | 30.6 (8.4) | 29.3 (7.3) | 31.9 (9.2) |
| Race, n (%) | Black | 2561 (51.2) | 1146 (45.8) | 1415 (56.6) |
| | White | 1895 (37.9) | 1044 (41.8) | 851 (34.0) |
| | Other | 544 (10.9) | 310 (12.4) | 234 (9.4) |

Training and Evaluation

All models were trained using PyTorch and PyTorch Lightning frameworks on an NVIDIA L40S GPU with a batch size of 16. To prevent overfitting, we employed an early stopping mechanism that halted training if validation performance did not improve (loss delta threshold of 1e-4) for five consecutive epochs.

We assessed model performance using standard metrics: Area Under the Receiver Operating Characteristic curve (AUROC) for classification tasks and Dice coefficient for segmentation tasks. To probe the models' limits, we conducted a subgroup analysis on challenging cases of small pneumothorax, defined as those in the first quartile of mask volume. We stratified this analysis by the presence or absence of a chest tube to investigate whether models were learning clinically relevant features or relying on confounding shortcuts (24).

For statistical comparisons between different models, we used the Friedman test, followed by the Nemenyi post-hoc test for pairwise comparisons. To evaluate the effect of confounders in the subgroup analysis, the Mann-Whitney U test was used to compare model performance on cases with a chest tube versus those without.

# RESULTS

We evaluated the quality and adaptability of image encoder features from various medical FMs across classification (**Figure 2**) and segmentation (**Figure 3**) tasks, comparing performance using a linear probing approach to assess inherent feature quality and a fine-tuning approach to assess feature adaptability.

Classification Performance

*pneumothorax*

For the pneumothorax classification task, linear probing of the medical FMs achieved AUROC values ranging from 0.877 (BiomedCLIP) to 0.964 (Ark+), which surpassed the general models DINOv2 (0.870) and SigLIP2 (0.871). The statistical significance heatmap (**Figure 1b**) confirmed that DINOv2 and SigLIP2 showed significant underperformance compared to Ark+ and MedImageInsights. Notably, Ark+ significantly outperformed BiomedCLIP. All FMs' linear probing scores did not show any significant difference to SegFormer's performance (0.89). Fine-tuning yielded statistically significant improvements for models with lower initial linear probing scores: BiomedCLIP (0.877 to 0.879, p=0.025), RAD-DINO (0.915 to 0.924, p=0.025), and SigLIP2 (0.871 to 0.895, p=0.025). Models with strong linear probing performance—Ark+, CheXagent, and MedImageInsights—showed only non-significant improvements. Ark+ was the only model to significantly outperform Segformer (0.887).

*cardiomegaly*

For cardiomegaly prediction, we similarly observed that medical models reached 0.892 – 0.945 AUROC, compared to 0.842–0.866 for general models. Statistical analysis revealed DINOv2 was significantly inferior to Ark+, CheXagent, and MedImageInsights.

Three models demonstrated significant fine-tuning benefits: BiomedCLIP (0.892 to 0.908, p=0.025), DINOv2 (0.842 to 0.872, p=0.025), and SigLIP2 (0.866 to 0.919, p=0.025). MedImageInsights showed a slight but performance decrease (0.945 to 0.937, p=0.025) after fine-tuning. The traditional supervised baseline, SegFormer, was significantly outperformed by Ark+ and CheXagent after fine-tuning.

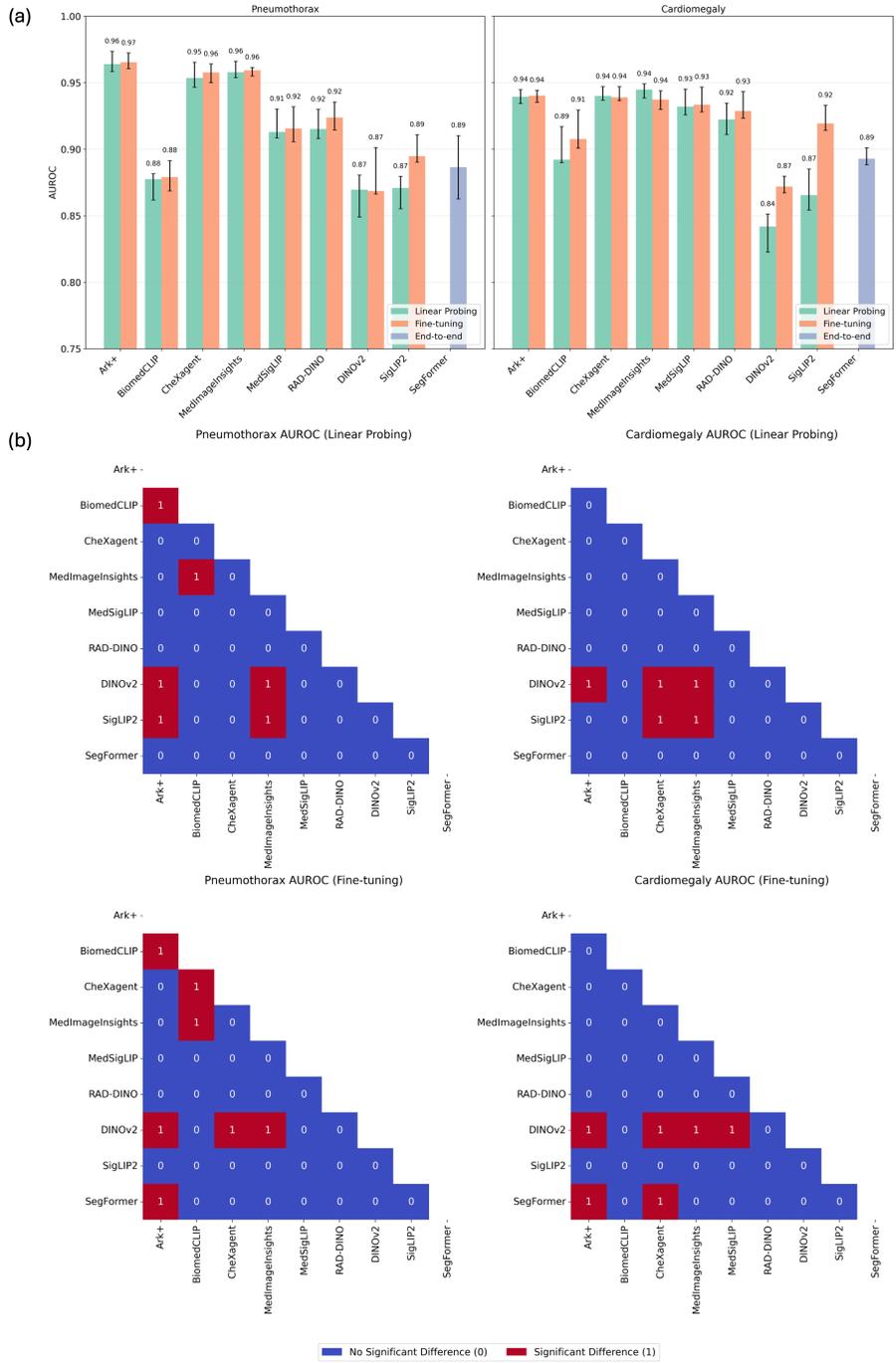

**Figure 2.** Classification Performance and Statistical Analysis for pneumothorax and cardiomegaly. (a) The bar charts display the Area Under the Receiver Operating Characteristic curve (AUROC) for each FM under both linear probing and fine-tuning evaluations. Error bars denote the 95% confidence interval. (b) The heatmaps present pairwise statistical comparisons of model performance for both linear probing and fine-tuning. A red square (1) indicates a statistically significant difference in performance ($p<0.05$), while a blue square (0) signifies no significant difference. p values are shown in **Figure S3**.

Segmentation Performance

The analysis of effective patch sizes (16 vs. 32) within hierarchical architectures (Ark+, MedImageInsights) consistently showed that the smaller patch size of 16 yields superior linear probing performance for segmentation tasks. For example, in pneumothorax segmentation, Ark+ with 16x16 patches achieved a 0.433 Dice score compared to 0.380 Dice with 32x32 patches. Consequently, all comparisons with other models were made using the results obtained from the 16×16 patch configuration. Intersection over Union (IoU) was also measured, showing a similar trend to the Dice scores (**Figure S2**).

*Pneumothorax Segmentation*

For segmenting small pneumothorax (ground truth pneumothorax mask within the first quartile) (**Figure 4**), performance was uniformly poor across all FMs with linear probing, ranging from 0.219 (BiomedCLIP) to 0.433 Dice (Ark+). BiomedCLIP's performance was significantly inferior to Ark+ and RAD-DINO.

Fine-tuning was critical, yielding substantial and statistically significant gains for seven of eight FMs; only Ark+ showed non-significant improvement. The supervised baseline SegFormer achieved the highest overall performance at 0.478 Dice, but MedImageInsights closely matched this performance (0.476 Dice) after fine-tuning. After fine-tuning, MedImageInsights and SegFormer significantly outperformed BiomedCLIP, MedSigLIP, and DINOv2.

*Cardiac Segmentation*

For segmenting the heart (**Figure 5**), a distinct anatomical structure, all FMs achieved strong initial performance with linear probing (0.890 to 0.942 Dice, p=0.025). The medical models RAD-DINO, MedImageInsights, and the supervised baseline SegFormer significantly outperformed the general-domain models (DINOv2 and SigLIP2) in this initial phase.

Fine-tuning was highly effective, nearly all models (7/8) showing significant improvements. All models converged to a high-performance range of 0.908–0.961 Dice. MedImageInsights achieved the highest Dice score (0.961), matching the performance of the end-to-end supervised baseline SegFormer.

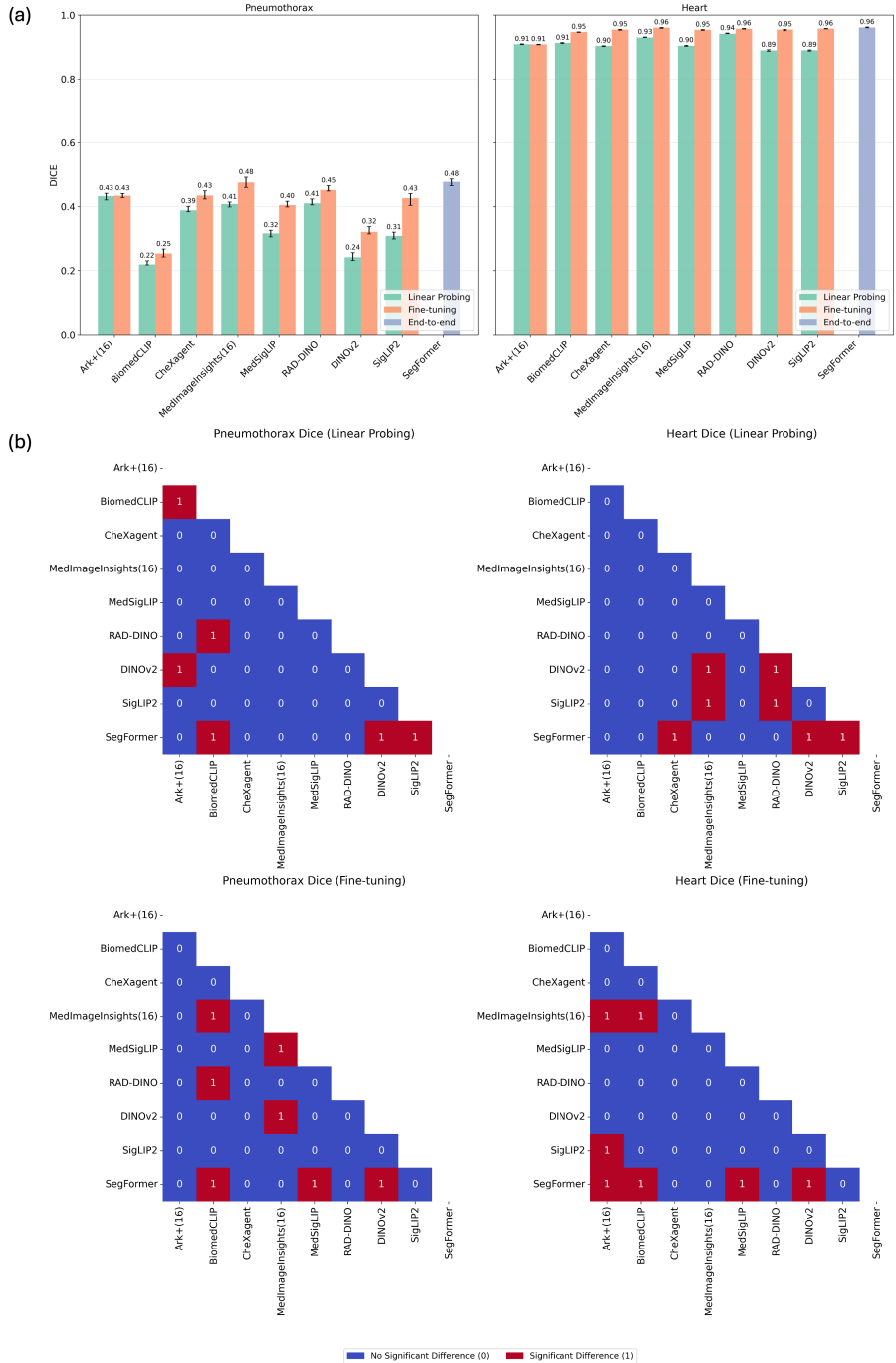

**Figure 3.** Segmentation Performance and Statistical Analysis for pneumothorax and cardiomegaly. (a) The bar charts display the dice scores for each FM under both linear probing and fine-tuning evaluations. Error bars denote the 95% confidence interval. (b) The heatmaps present pairwise statistical comparisons of model performance for both linear probing and fine-tuning. A red square (1) indicates a statistically significant difference in performance ($p<0.05$), while a blue square (0) signifies no significant difference. p values are shown in **Figure S4**.

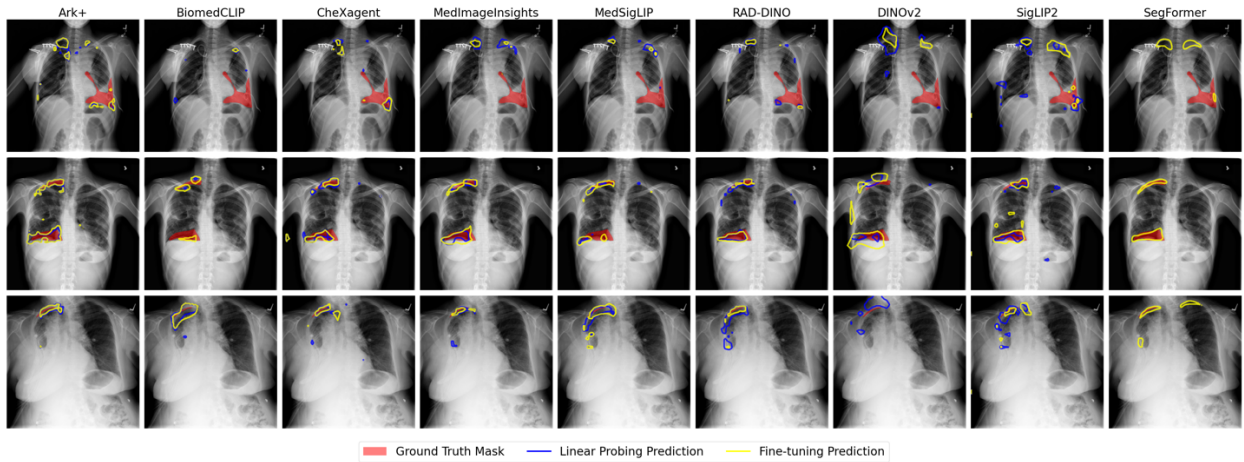

**Figure 4.** pneumothorax Segmentation Performance of FMs. This comparison illustrates the quality and adaptability of the learned spatial features from each FM for this dense prediction task.

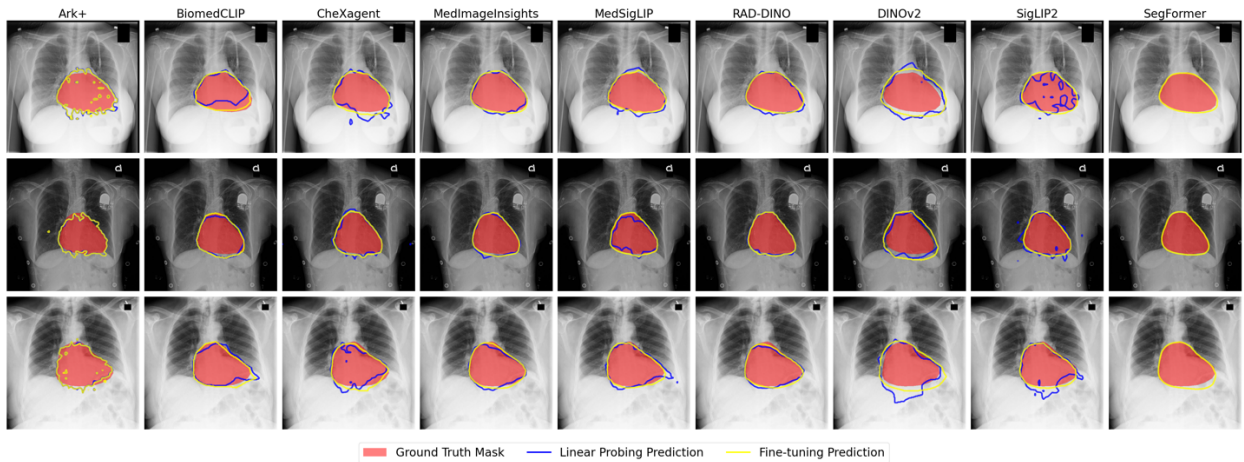

**Figure 5.** Cardiac Segmentation Performance of FMs. This comparison illustrates the quality and adaptability of the learned spatial features from each FM for this dense prediction task.

Subgroup Analysis on Small pneumothorax

We performed a subgroup analysis focusing on cases of small pneumothorax, dividing them based on whether patients had chest tubes inserted.

The results revealed a dramatic difference in performance between the two subgroups, particularly for classification (**Figure 6** and **Table S2**). On images with a chest tube present, most models demonstrated high classification performance. For example, MedImageInsights and MedSigLIP achieved high linear probing sensitivity of 0.967 and 0.956, respectively. In contrast, on images without a chest tube, classification performance plummeted to levels no better than a random guessing for all models in the linear probing phase. While fine-tuning offered some significant improvements, such as for MedImageInsights (0.720 sensitivity) and Ark+ (0.586 sensitivity), performance remained modest overall.

Segmentation performance (**Figure 7**), however, was uniformly poor across both subgroups, regardless of the presence of a chest tube. After fine-tuning, although all performances improved significantly, Dice scores for segmenting small pneumothoraces remained low, with the top-performing supervised baseline (SegFormer) achieving a Dice score of only 0.308 on cases without a chest tube and 0.248 on cases with one.

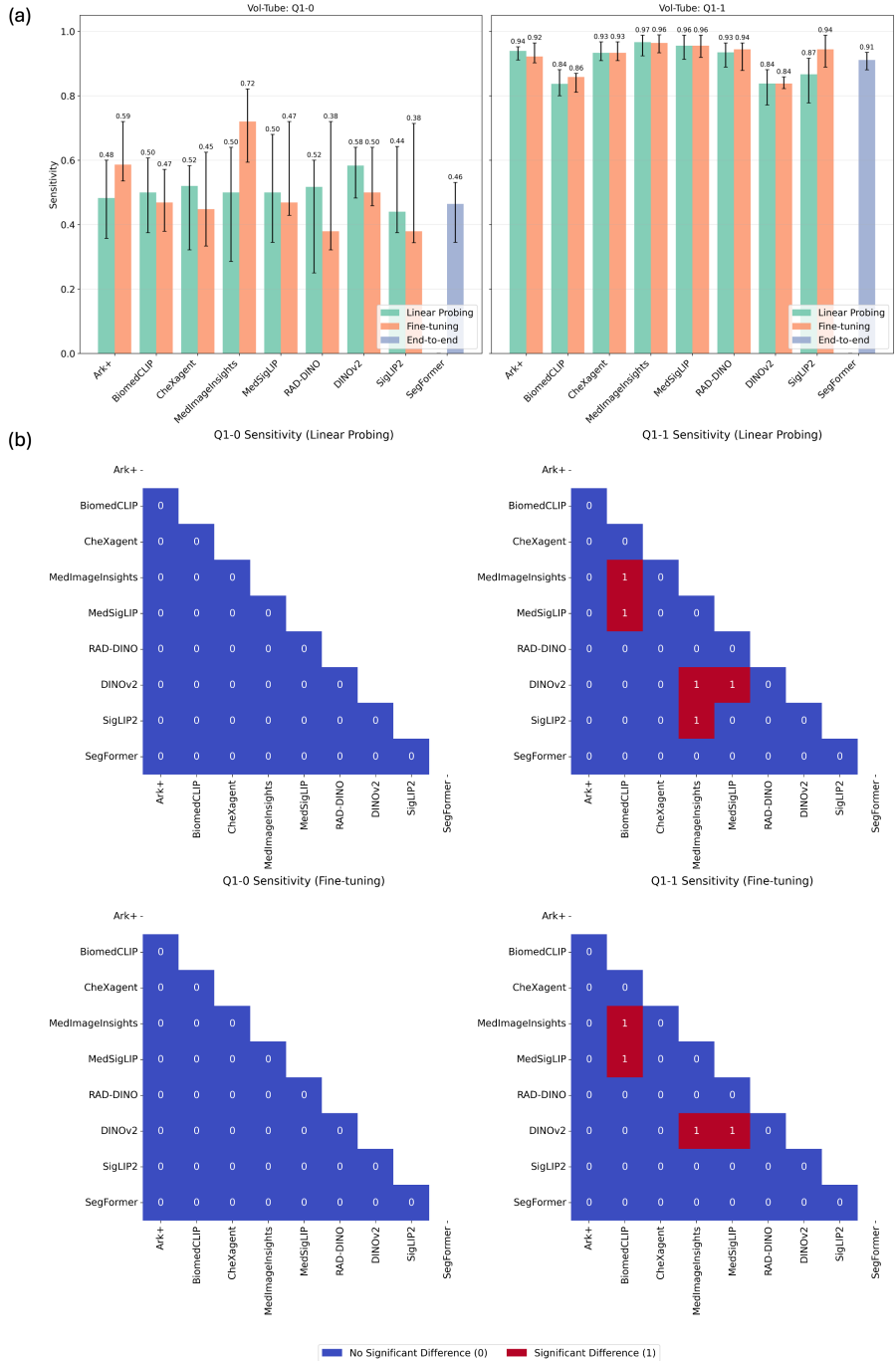

**Figure 6.** Subgroup Analysis of Classification Performance on Challenging Small pneumothorax Cases with and without Chest Tube. Chest Tube is a confounding factor in pneumothorax detection. (a) The bar charts display the sensitivity for each FM under both linear probing and fine-tuning evaluations. Error bars denote the 95% confidence interval. (b) The heatmaps present pairwise statistical comparisons of model performance for both linear probing and fine-tuning. A red square (1) indicates a statistically significant difference in performance ($p<0.05$), while a blue square (0) signifies no significant difference. p values are shown in **Figure S5**.

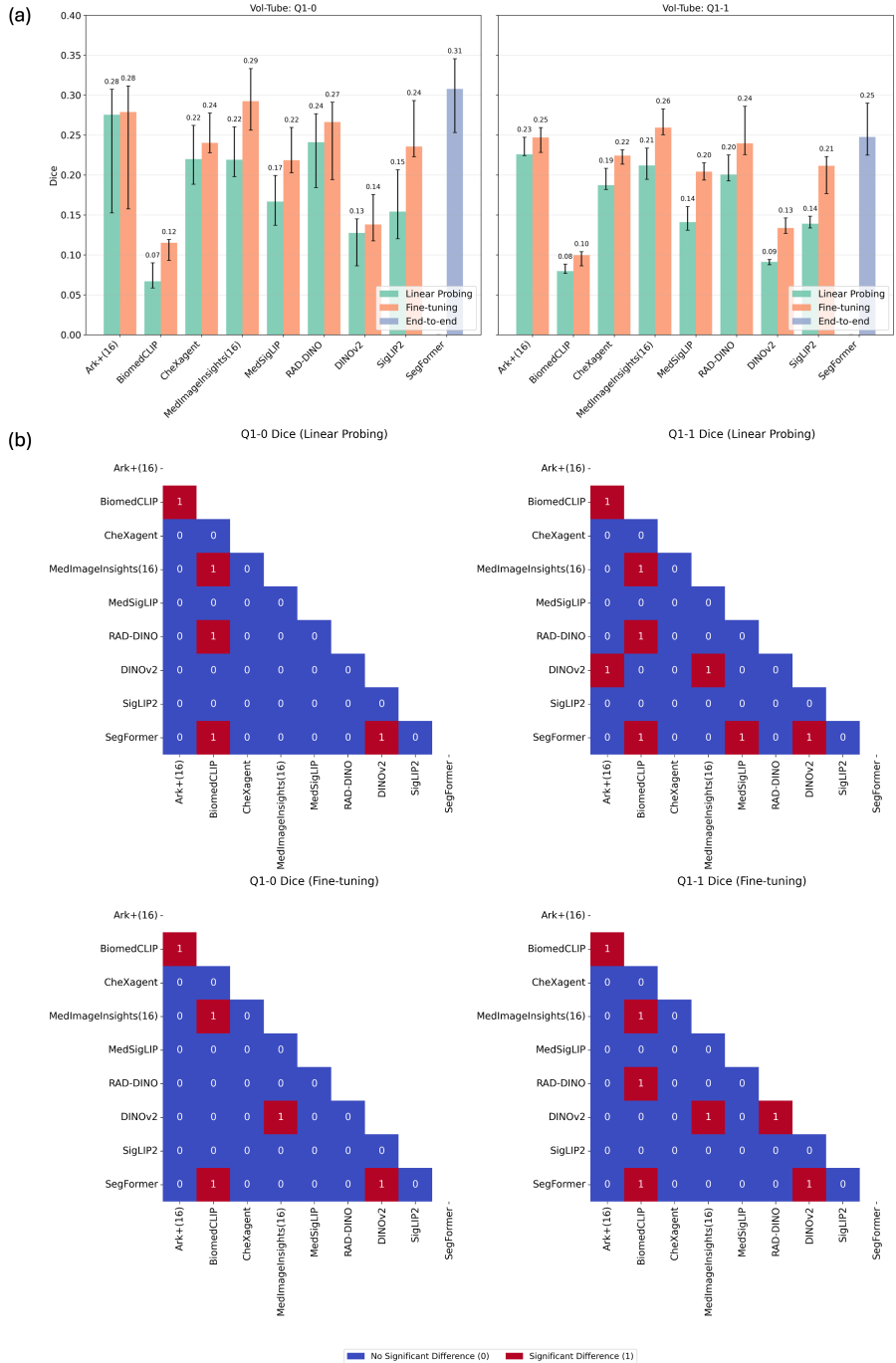

**Figure 7.** Subgroup Analysis of Segmentation Performance on Challenging Small pneumothorax Cases with and without Chest Tube. Chest Tube is a confounding factor in pneumothorax detection. (a) The bar charts display the Dice score for each FM under both linear probing and fine-tuning evaluations. Error bars denote the 95% confidence interval. (b) The heatmaps present pairwise statistical comparisons of model performance for both linear probing and fine-tuning. A red square (1) indicates a statistically significant difference in performance ($p<0.05$), while a blue square (0) signifies no significant difference. p values are shown in **Figure S6**.

# DISCUSSION

## Domain Pretraining, Feature Quality, and Adaptability

The consistent and statistically significant superiority of medical FMs over general-domain models in linear probing establishes the value of domain-specific pretraining for initial feature quality. For example, in pneumothorax classification, medical FMs like Ark+ (0.964 AUROC) and MedImageInsights (0.958 AUROC) achieved high performance with linear probing (25), significantly outperforming general models like DINOv2 (0.870 AUROC) and SigLIP2 (0.871 AUROC). This domain benefit was also evident for cardiomegaly classification, where the performance gap reached up to 0.10 AUROC in the linear probing phase between the top performing medical FMs and the general models. This hierarchy demonstrates that medical pretraining effectively encodes task-appropriate features, making the models robust prior to any fine-tuning.

The linear probing results effectively measure inherent feature quality and predict the required adaptability investment. Models with the highest initial performance (e.g., Ark+, CheXagent) showed non-significant fine-tuning improvements across most classification tasks, indicating their features were already task-appropriate. Conversely, models with weaker initial performance consistently showed significant fine-tuning benefits. While SigLIP2 showed significant gains in all four tasks, DINOv2 showed significant gains in three of four tasks, necessitating a substantial representational shift to bridge the domain gap. This inverse relationship underscores that linear probing is a reliable proxy for feature quality: high initial scores mean features are robust and require minimal adaptation. Notably, MedImageInsights was a unique exception, showing a significant performance decrease from 0.945 to 0.937 AUROC for cardiomegaly after fine-tuning.

## Task-Dependent Feature Utility and Adaptation Strategies

The required level of feature adaptation is highly task-dependent, especially when contrasting classification and segmentation, as well as simple anatomy versus complex pathologies. For segmenting pneumothorax, linear probing performance was uniformly poor across all FMs, ranging from 0.219 (BiomedCLIP) to 0.433 Dice (Ark+). This inability to localize a complex, subtle pathology suggests that medical pretraining does not automatically encode the necessary fine-grained spatial details. This limitation made fine-tuning critical, yielding substantial and statistically significant gains for seven of eight FMs (**Table S1**). The supervised baseline SegFormer achieved the highest overall performance at 0.478 Dice, but MedImageInsights closely matched this performance (0.476 Dice) after fine-tuning.

In contrast, segmenting the heart anatomy, which is more salient, showed strong initial linear probing performance (0.890 to 0.942 Dice). The general models (DINOv2, SigLIP2) required substantial, significant improvements from fine-tuning (e.g., 0.890 Dice improving to >0.95 Dice), aligning with the required adaptation observed in general FMs for classification. All FMs showed significant fine-tuning improvements for this task, with performance converging to a high-performance ceiling (0.947−0.961 Dice). This dichotomy demonstrates that not all segmentation tasks are equal: the models' pre-trained features are more easily adapted for segmenting salient anatomical structures (e.g. heart) than subtle, complex pathologies (e.g. pneumothorax).

Comparison to End-to-End Training

While fine-tuned FMs generally surpassed the supervised baseline for classification (e.g., Ark+ and CheXagent significantly outperformed SegFormer for cardiomegaly), SegFormer achieved highly competitive or superior results in segmentation. For segmentation tasks, end-to-end training matched the best FM for cardiac segmentation (0.961 Dice) and achieved the highest overall pneumothorax segmentation score (0.478 Dice), closely followed by MedImageInsights (0.476 Dice). This suggests that for tasks demanding high spatial precision, direct end-to-end optimization of an appropriate architecture may be equally effective as adapting pretrained features.

Subgroup Analysis: The Challenge of Small pneumothorax

To further probe the models' capabilities and limitations, we conducted a subgroup analysis on challenging cases of small pneumothorax, stratifying the data by the presence or absence of a chest tube—a known clinical confounder (24). This allowed us to assess model robustness and understand the specific visual features the models rely on. The distinct contrast between classification and segmentation performance on the subgroup analysis highlights a fundamental difference in the features required for each task. Classification primarily answers the question, "What is in this image?", while segmentation must answer, "Where is it precisely?"

The models' high classification sensitivity on images containing a chest tube, paired with their failure on images without one, strongly suggests they learned a powerful shortcut. Instead of identifying the subtle pleural line of the pneumothorax itself, the models learned to associate the highly conspicuous presence of a chest tube with the diagnosis. While statistically valid for classification, this correlational feature is a proxy and not the actual pathology. The models succeeded by identifying an obvious, related object rather than the clinical finding of interest.

This reliance on shortcuts is sufficient for classification, which can succeed using high-level, correlational patterns. However, it is inadequate for segmentation, which demands a precise, low-level spatial understanding of the pathology's boundary. The consistently poor segmentation scores—even on images with chest tubes where classification was successful—suggest that the models' pretrained features might not inherently contain all the fine-grained information needed to delineate the pleural line.

Model Architectures and Pre-traninig Paradigms for Vision Encoders

Empirical results suggest a distinct performance advantage for hierarchical architectures like Ark+ and MedImageInsights. Their ability to capture multi-scale features appears inherently well-suited for medical imaging, which requires processing both fine-grained pathological details and broader anatomical context. This is further supported by our finding that smaller effective patch sizes (16×16 versus 32×32) consistently yield superior segmentation performance, underscoring the importance of preserving high-resolution spatial information.

Beyond architecture, the pre-training paradigm is a key determinant of feature quality. Models like RAD-DINO, pre-trained using only image data without text alignment, demonstrated competitive performance, particularly in segmentation tasks. This suggests that self-supervised visual learning can produce granular representations beneficial for detailed spatial understanding. Similarly, Ark+, trained with direct image-label supervision, also achieved strong results across diverse tasks. Together, these models indicate that expensive image-text alignment is not a prerequisite for developing powerful vision encoders intended for downstream adaptation, especially when zero-shot text-based inference is not the primary goal.

In contrast, BiomedCLIP showed comparatively weaker performance. This could be attributed to its pre-training on curated, prototypical images from academic publications, which may not capture real-world data variability. Furthermore, its low input resolution (224×224) and coarse patch embeddings (14×14) may lead to a critical loss of detail when processing high-resolution radiological images.

Limitations and Practical Challenges

Several limitations should be considered. Our evaluation was confined to chest radiographs and two specific clinical findings; future work should explore generalizability across different imaging modalities and pathologies. The performance of each model is also intrinsically linked to the quality and scale of its pre-training dataset, which can influence adaptability. Specific to our study, we utilized a third-party implementation of MedImageInsights, which may not perfectly reflect its intended performance, and noted that CheXagent had prior exposure to the pneumothorax dataset during its pre-training. Finally, our use of five-fold cross-validation may limit the statistical power to detect more subtle performance differences between models.

Beyond these experimental limitations, we encountered significant practical hurdles that challenge the adoption of FMs in research. The process of extracting embeddings is often complicated by poor documentation, particularly regarding essential image preprocessing steps (e.g., bit depth, intensity scaling). This frequently requires users to inspect source code to determine the correct input format—a task that is impossible for closed-source models. Furthermore, the need to extract different types of embeddings for different tasks (e.g., global embeddings for classification, patch embeddings for segmentation) adds a layer of complexity and restricts flexibility, demanding an in-depth understanding of each model's internal architecture. These challenges highlight a critical need for standardized practices to streamline the use of FMs in downstream applications.

# CONCLUSION

Our findings demonstrate that domain-specific pre-training on medical images provides a substantial advantage, equipping FMs with higher-quality and more adaptable features than their general-purpose counterparts. However, the utility of these features is highly task-dependent. While pre-trained representations are often sufficient for global classification tasks, they frequently fall short in dense prediction tasks like segmentation, which require significant fine-tuning to achieve strong performance. This suggests that current medical FMs do not yet fully capture the fine-grained spatial information necessary for precise localization.

Architectural choices, such as the multi-scale design of hierarchical models, appear to provide an inherent advantage for the unique demands of medical imaging. Moving forward, to maximize efficiency and clinical utility, the field should focus on developing FMs that are architecturally suited for both global disease detection and dense prediction. Finally, our results demonstrate that image-text alignment may not be necessary to achieve good performance when adapting a FM to downstream tasks, especially where zero-shot inference or text-based interaction is not required.

# DECLARATIONS

Ethics Approval and Consent to Participate

This study was approved by the institutional review board of Emory University (STUDY00002276: Quality and Informatics Protocol) and informed consent was waived. The manuscript and the results were presented in a way that the patients cannot be identified.

Competing Interests

There is no conflict of interest for all authors.

Funding/Support

This study was supported by the AI Image Extraction Core, an Emory Integrated Core Facility. Dr. Gichoya is a 2022 Robert Wood Johnson Foundation Harold Amos Medical Faculty Development Program and declares support from Lacuna Fund (#67), Gordon and Betty Moore Foundation, NIH (NIBIB) MIDRC grant under contracts 75N92020C00008 and 75N92020C00021, NHLBI Award Number R01HL167811 and NIH common fund award 1R25OD039834-01.

Author's Contributions

All authors made a significant contribution to the work reported, whether that is in the conception, study design, execution, acquisition of data, analysis and interpretation, or in all these areas; took part in drafting, revising or critically reviewing the article; gave final approval of the version to be published; have agreed on the journal to which the article has been submitted; and agree to be accountable for all aspects of the work.

Data Statement

The SIIM-ACR pneumothorax dataset is publicly available through the Kaggle competition platform at https://www.kaggle.com/competitions/siim-acr-pneumothorax-segmentation.

The institutional chest X-ray dataset used for cardiomegaly analysis is not publicly available as it may contain protected health information. Please contact the corresponding author for requests regarding access to this data.

Declaration Of Generative AI And AI-Assisted Technologies

During the preparation of this work the author(s) used Gemini, a large language model from Google, in order to improve the language and correct grammar. After using this tool/service, the author(s) reviewed and edited the content as needed and take(s) full responsibility for the content of the published article.

# SUPPLEMENTARY MATERIAL

Linear Probing for Classification

The linear probing classifier consists of a single linear layer that takes high-dimensional CLS embeddings and projects them to the binary classes. The model uses binary cross-entropy with logits loss for binary classification tasks. It's designed to be used for evaluating the quality of pre-trained CLS embeddings by measuring how well they can be linearly separated for downstream tasks.

Linear Probing for Segmentation

The linear probing segmentation model leverages patch embeddings from a pre-trained vision encoder. It features a single 1×1 convolutional layer that efficiently projects high-dimensional feature maps (e.g. 768 channels at 37×37 resolution for RAD-DINO) to a target class (e.g. pneumothorax), followed by bilinear upsampling to restore the original image resolution (e.g. 518×518 for RAD-DINO). This minimalist architecture helps determine whether rich semantic information already exists within the patch embeddings (1). The model functions as both a baseline and a reference point for ablation studies when evaluating more complex segmentation approaches, and utilizes binary cross-entropy with logits loss for training.

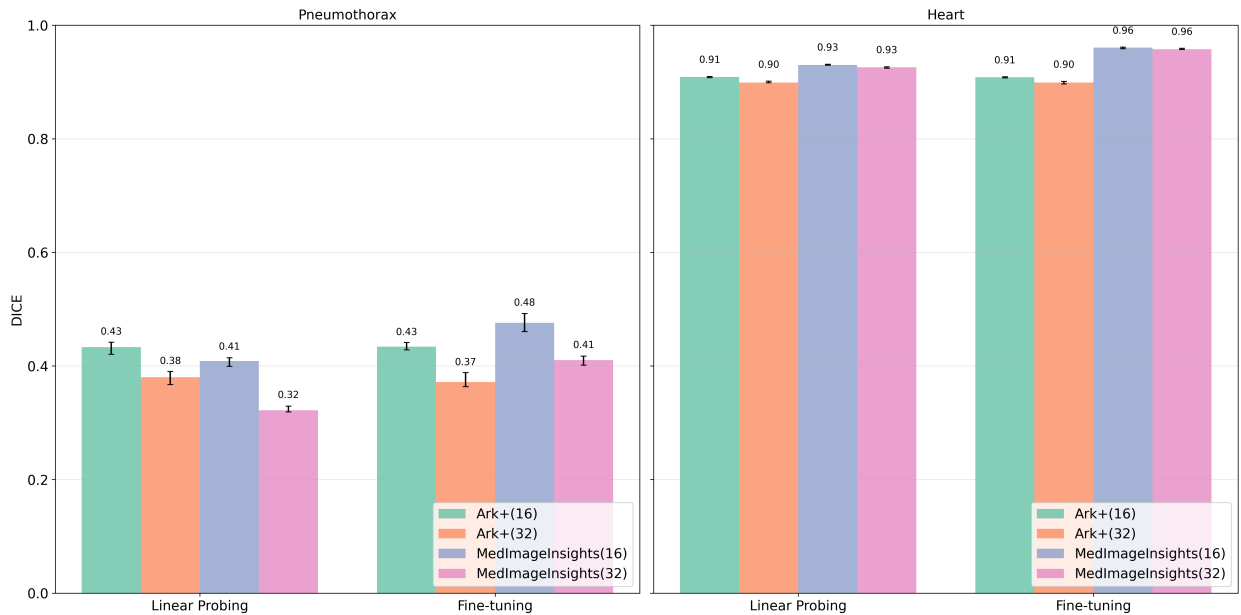

**Figure S8.** Impact of Effective Patch Size on Downstream Segmentation Performance. This figure illustrates the segmentation performance of features extracted from two hierarchical models, Ark+ and MedImageInsights, at different effective patch sizes (16 vs. 32). On the challenging Pneumothorax task (left), the smaller patch size of 16 yields superior results for both linear probing and fine-tuning. For the Heart segmentation task (right), all configurations achieve

excellent performance (DICE > 0.90. Similarly, the smaller patch size of 16 performs either marginally better or equally well.

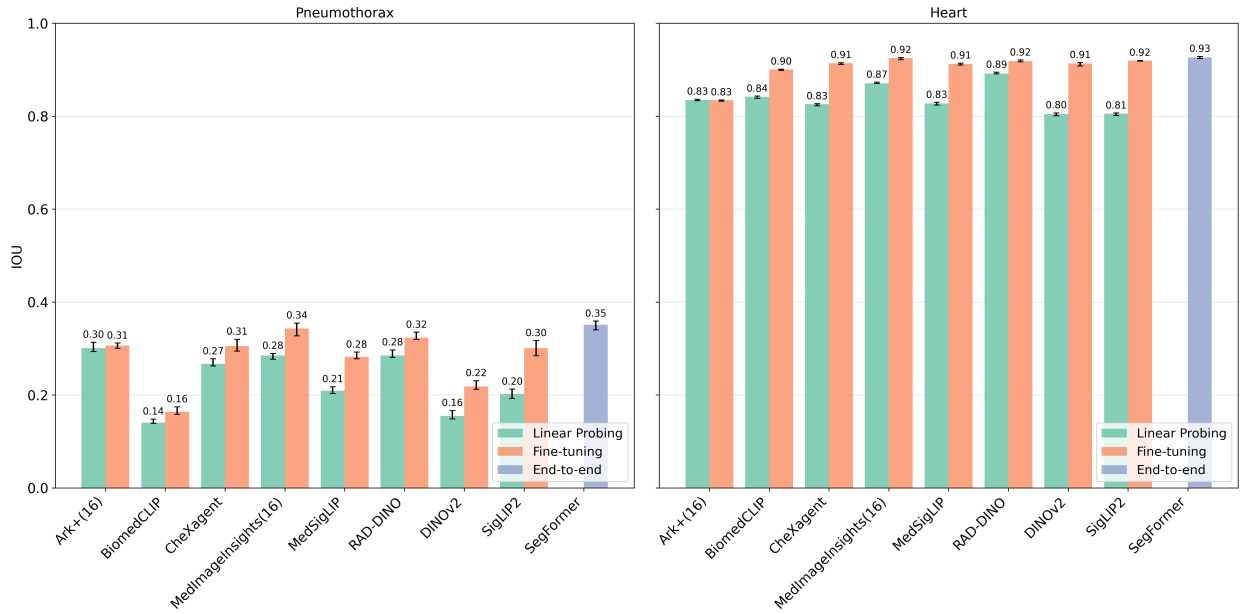

**Figure S9.** Segmentation Performance and Statistical Analysis for Pneumothorax and Cardiomegaly. (a) The bar charts display the IoU scores for each FM under both linear probing and fine-tuning evaluations. Error bars denote the 95% confidence interval.

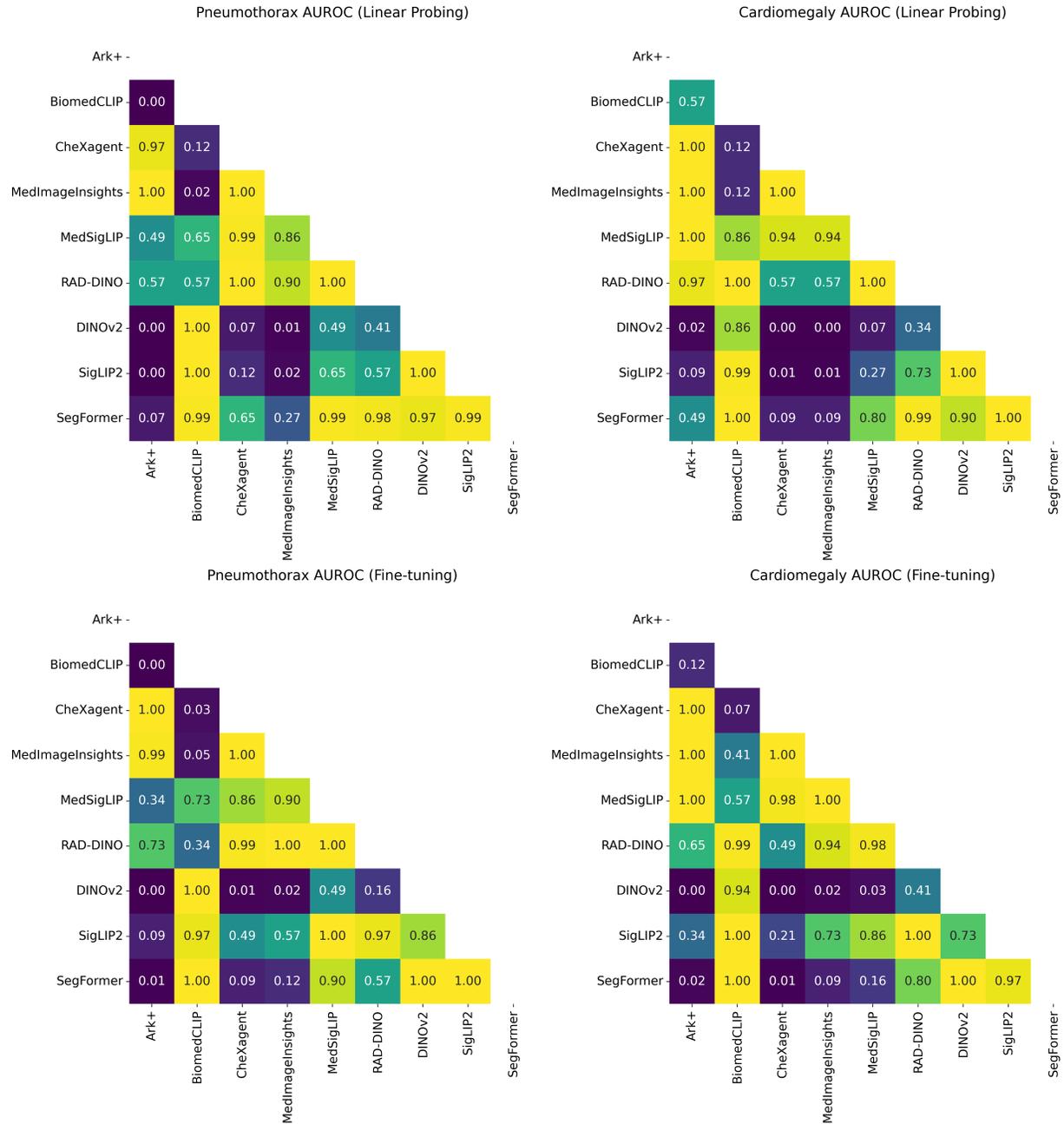

**Figure S10.** The heatmaps present p values of pairwise statistical comparisons of model performance in classification for both linear probing and fine-tuning.

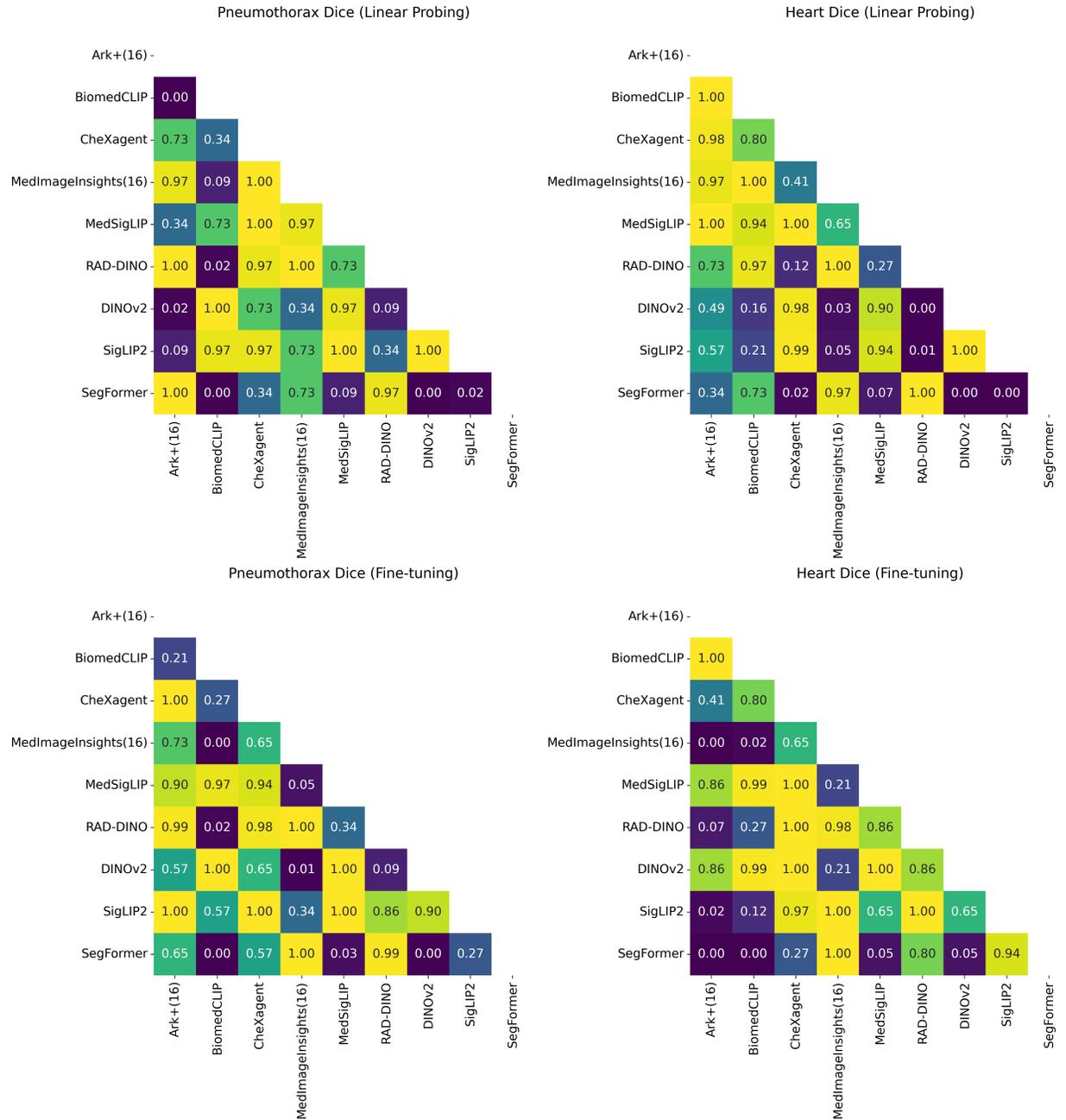

**Figure S11.** The heatmaps present p values of pairwise statistical comparisons of model performance in segmentation for both linear probing and fine-tuning.

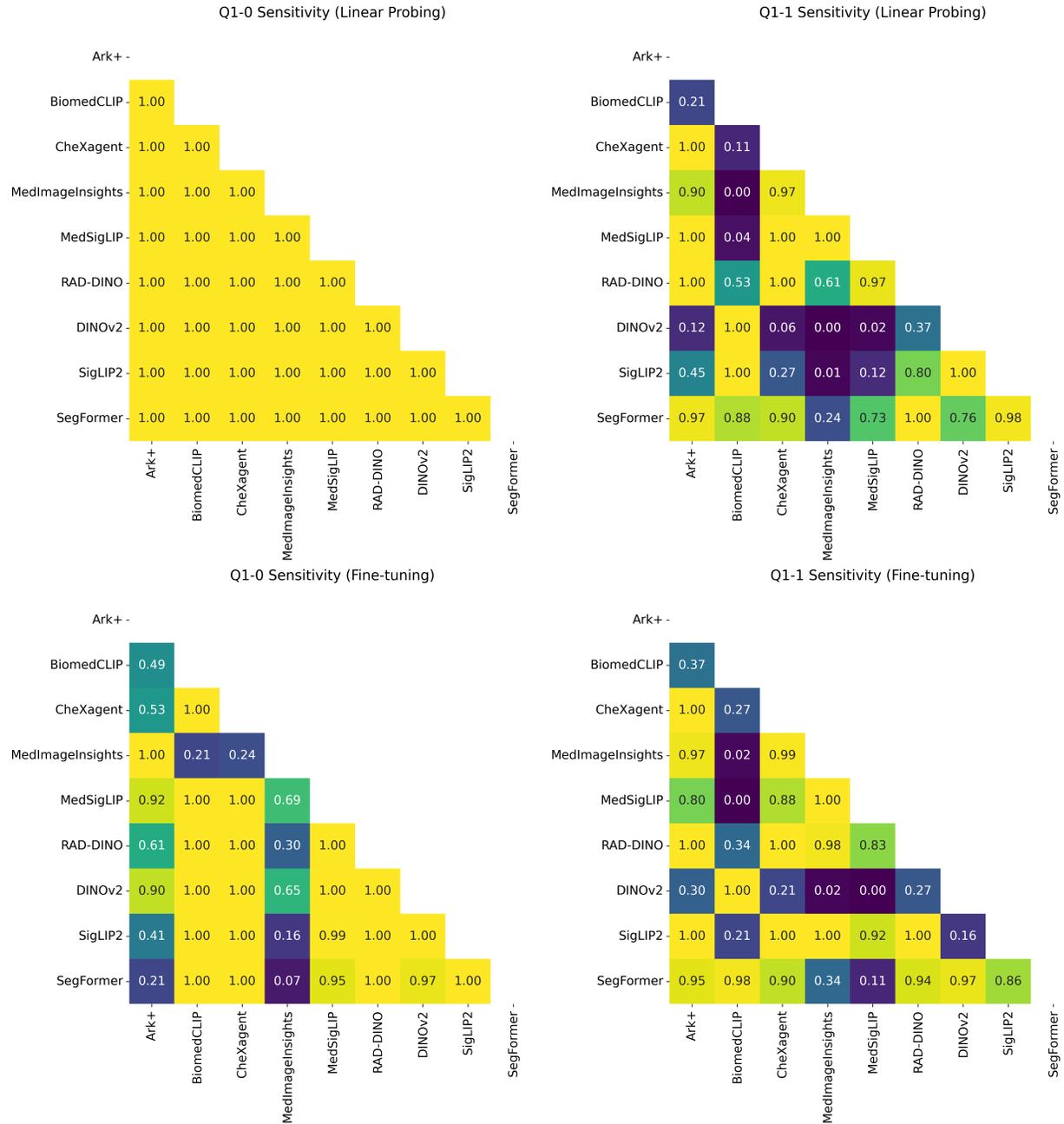

**Figure S12.** The heatmaps present p values of pairwise statistical comparisons of model performance in subgroup analysis of classification for both linear probing and fine-tuning.

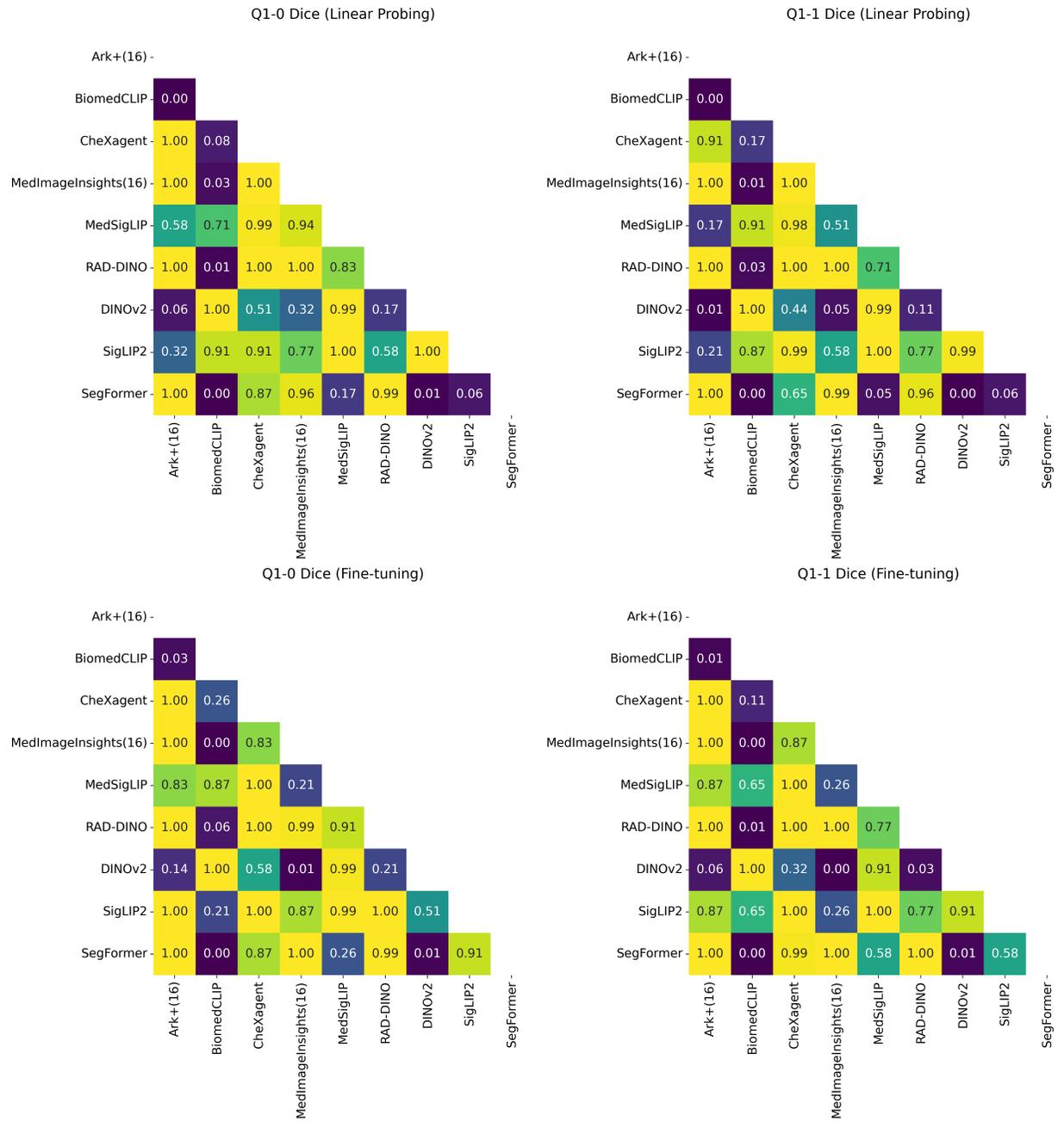

**Figure S13.** The heatmaps present p values of pairwise statistical comparisons of model performance in subgroup analysis of segmentation for both linear probing and fine-tuning.

**Table S4.** A comparison of model performance on classification (AUROC) and segmentation (Dice) tasks. Metrics are reported as median with 95% confidence intervals. p-values indicate the statistical significance of the difference between linear probing and fine-tuning.

| Task | Condition | Model | Linear Probing | Fine-tuning | p | Significance |
|---|---|---|---|---|---|---|
| **Classification** | Pneumothorax | Ark+ | 0.964 (0.958-0.974) | 0.965 (0.961-0.972) | 0.655 | False |
| | | BiomedCLIP | 0.877 (0.862-0.881) | 0.879 (0.869-0.891) | 0.025 | True |
| | | CheXagent | 0.954 (0.947-0.965) | 0.958 (0.950-0.964) | 0.180 | False |
| | | MedImageInsights | 0.958 (0.954-0.966) | 0.959 (0.955-0.961) | 0.655 | False |
| | | MedSigLIP | 0.913 (0.908-0.930) | 0.916 (0.906-0.932) | 0.655 | False |
| | | RAD-DINO | 0.915 (0.908-0.930) | 0.924 (0.914-0.936) | 0.025 | True |
| | | DINOv2 | 0.870 (0.849-0.881) | 0.869 (0.866-0.901) | 0.655 | False |
| | | SigLIP2 | 0.871 (0.855-0.880) | 0.895 (0.890-0.911) | 0.025 | True |
| | | SegFormer | | 0.887 (0.863-0.910)* | | |
| | Cardiomegaly | Ark+ | 0.939 (0.934-0.945) | 0.940 (0.935-0.944) | 0.180 | False |
| | | BiomedCLIP | 0.892 (0.890-0.917) | 0.908 (0.901-0.930) | 0.025 | True |
| | | CheXagent | 0.940 (0.937-0.947) | 0.939 (0.936-0.947) | 0.655 | False |
| | | MedImageInsights | 0.945 (0.938-0.949) | 0.937 (0.930-0.944) | 0.025 | True |
| | | MedSigLIP | 0.932 (0.926-0.945) | 0.933 (0.928-0.947) | 0.655 | False |
| | | RAD-DINO | 0.922 (0.911-0.935) | 0.929 (0.923-0.943) | 0.180 | False |
| | | DINOv2 | 0.842 (0.823-0.851) | 0.872 (0.867-0.880) | 0.025 | True |
| | | SigLIP2 | 0.866 (0.854-0.885) | 0.919 (0.914-0.933) | 0.025 | True |
| | | SegFormer | | 0.893 (0.888-0.901)* | | |
| **Segmentation** | Pneumothorax | Ark+(16) | 0.433 (0.421-0.442) | 0.434 (0.428-0.441) | 0.180 | False |
| | | BiomedCLIP | 0.219 (0.218-0.231) | 0.254 (0.243-0.267) | 0.025 | True |
| | | CheXagent | 0.388 (0.384-0.401) | 0.435 (0.424-0.449) | 0.025 | True |
| | | MedImageInsights(16) | 0.409 (0.399-0.415) | 0.476 (0.460-0.492) | 0.025 | True |
| | | MedSigLIP | 0.316 (0.305-0.326) | 0.405 (0.399-0.417) | 0.025 | True |
| | | RAD-DINO | 0.410 (0.406-0.424) | 0.452 (0.449-0.466) | 0.025 | True |
| | | DINOv2 | 0.242 (0.232-0.256) | 0.321 (0.315-0.338) | 0.025 | True |
| | | SigLIP2 | 0.308 (0.299-0.320) | 0.426 (0.404-0.441) | 0.025 | True |
| | | SegFormer | | 0.478 (0.466-0.487)* | | |
| | Cardiac Segmentation | Ark+(16) | 0.909 (0.908-0.910) | 0.908 (0.907-0.909) | 0.025 | True |
| | | BiomedCLIP | 0.913 (0.911-0.914) | 0.947 (0.946-0.947) | 0.025 | True |
| | | CheXagent | 0.903 (0.901-0.904) | 0.955 (0.953-0.955) | 0.025 | True |
| | | MedImageInsights(16) | 0.930 (0.930-0.931) | 0.961 (0.959-0.961) | 0.025 | True |
| | | MedSigLIP | 0.904 (0.902-0.906) | 0.954 (0.952-0.955) | 0.025 | True |
| | | RAD-DINO | 0.942 (0.941-0.944) | 0.957 (0.956-0.958) | 0.025 | True |
| | | DINOv2 | 0.890 (0.887-0.891) | 0.954 (0.951-0.955) | 0.025 | True |
| | | SigLIP2 | 0.890 (0.887-0.891) | 0.958 (0.957-0.958) | 0.025 | True |
| | | SegFormer | | 0.961 (0.960-0.963)* | | |

*SegFormer, a deep learning model trained end-to-end, served as the supervised baseline.

**Table S5.** Performance on the small pneumothorax subgroup, stratified by chest tube presence. Metrics for classification (AUROC) and segmentation (Dice) are reported as median (95% CI). P-values indicate the significance of performance changes after fine-tuning. <u>Underlined values</u> denote a significant difference between the presence and absence of a chest tube.

| Task | Chest Tube | Model | Linear Probing | Fine-tuning | $p$ | Significance |
|---|---|---|---|---|---|---|
| Classification | Absent | Ark+ | 0.483 (0.357-0.600) | 0.586 (0.536-0.720) | 0.025 | True |
| | | BiomedCLIP | 0.500 (0.375-0.607) | 0.469 (0.379-0.571) | 0.655 | False |
| | | CheXagent | 0.520 (0.321-0.583) | 0.448 (0.333-0.625) | 0.180 | False |
| | | DINOv2 | 0.583 (0.483-0.640) | 0.500 (0.458-0.640) | 0.317 | False |
| | | MedImageInsights | 0.500 (0.286-0.640) | 0.720 (0.594-0.821) | 0.025 | True |
| | | MedSigLIP | 0.500 (0.345-0.680) | 0.469 (0.429-0.720) | 0.655 | False |
| | | RAD-DINO | 0.517 (0.250-0.600) | 0.379 (0.321-0.720) | 0.655 | False |
| | | SigLIP2 | 0.440 (0.375-0.643) | 0.379 (0.344-0.714) | 1.000 | False |
| | | SegFormer | | 0.464 (0.345-0.531)* | | |
| | Present | Ark+ | <u>0.939 (0.911-0.952)</u> | <u>0.922 (0.902-0.964)</u> | 0.655 | False |
| | | BiomedCLIP | <u>0.837 (0.800-0.881)</u> | <u>0.859 (0.811-0.870)</u> | 0.564 | False |
| | | CheXagent | <u>0.933 (0.909-0.967)</u> | <u>0.933 (0.909-0.967)</u> | 0.157 | False |
| | | DINOv2 | <u>0.838 (0.772-0.881)</u> | <u>0.838 (0.822-0.859)</u> | 1.000 | False |
| | | MedImageInsights | <u>0.967 (0.924-0.988)</u> | <u>0.964 (0.933-0.989)</u> | 0.180 | False |
| | | MedSigLIP | <u>0.956 (0.913-0.988)</u> | <u>0.956 (0.919-0.988)</u> | 0.157 | False |
| | | RAD-DINO | <u>0.935 (0.889-0.964)</u> | <u>0.944 (0.879-0.964)</u> | 0.564 | False |
| | | SigLIP2 | <u>0.867 (0.778-0.917)</u> | <u>0.944 (0.889-0.988)</u> | 0.025 | True |
| | | SegFormer | | 0.911 (0.880-0.935)* | | |
| Segmentation | Absent | Ark+(16) | 0.276 (0.153-0.307) | 0.279 (0.158-0.312) | 0.025 | True |
| | | BiomedCLIP | 0.067 (0.058-0.090) | 0.115 (0.093-0.119) | 0.025 | True |
| | | CheXagent | 0.220 (0.188-0.262) | 0.240 (0.228-0.278) | 0.025 | True |
| | | DINOv2 | 0.128 (0.086-0.145) | 0.138 (0.117-0.176) | 0.025 | True |
| | | MedImageInsights(16) | 0.219 (0.198-0.260) | 0.292 (0.256-0.333) | 0.025 | True |
| | | MedSigLIP | 0.167 (0.137-0.199) | 0.218 (0.203-0.259) | 0.025 | True |
| | | RAD-DINO | 0.241 (0.184-0.277) | 0.266 (0.194-0.291) | 0.025 | True |
| | | SigLIP2 | 0.154 (0.120-0.207) | 0.236 (0.223-0.293) | 0.025 | True |
| | | SegFormer | | 0.308 (0.253-0.346)* | | |
| | Present | Ark+(16) | 0.226 (0.224-0.247) | 0.247 (0.228-0.259) | 0.025 | True |
| | | BiomedCLIP | 0.080 (0.077-0.088) | 0.100 (0.086-0.104) | 0.025 | True |
| | | CheXagent | 0.187 (0.182-0.208) | 0.224 (0.214-0.232) | 0.025 | True |
| | | DINOv2 | 0.091 (0.088-0.094) | 0.134 (0.127-0.146) | 0.025 | True |
| | | MedImageInsights(16) | 0.212 (0.195-0.234) | 0.259 (0.250-0.283) | 0.025 | True |
| | | MedSigLIP | 0.141 (0.131-0.161) | 0.204 (0.194-0.215) | 0.025 | True |
| | | RAD-DINO | 0.201 (0.193-0.225) | 0.240 (0.226-0.286) | 0.025 | True |
| | | SigLIP2 | 0.139 (0.134-0.148) | 0.212 (0.177-0.223) | 0.025 | True |
| | | SegFormer | | 0.248 (0.225-0.290)* | | |

*SegFormer, a deep learning model trained end-to-end, served as the supervised baseline.